\documentclass[pdflatex,sn-basic]{sn-jnl}

\usepackage{cite}
\usepackage{natbib}
\usepackage{amsmath,amssymb,amsfonts}
\usepackage{algorithmic}
\usepackage{graphicx}
\usepackage{textcomp}
\usepackage{comment}

\usepackage{fancyhdr}

\usepackage{subfig}
\usepackage[utf8]{inputenc} 
\usepackage{graphicx} 
\usepackage{url}
\usepackage{hyperref}
\usepackage{xurl}  
\usepackage{cite}
\usepackage{float}  
\usepackage{amsmath} 
\usepackage{amssymb} 

\usepackage[parfill]{parskip} 

\usepackage{verbatim}
\usepackage{booktabs} 
\usepackage{paralist} 
\usepackage{subfig} 

\begin{document}

\title{Adversarial Robustness of Vision in Open Foundation Models}


\author*[1]{\fnm{Jonathan} \sur{Fox}}

\author[1]{\fnm{William J} \sur{Buchanan}}

\author[1]{\fnm{Pavlos} \sur{Papadopoulos}}

\affil*[1]{\orgdiv{Blockpass ID Lab}, \orgname{Edinburgh Napier University}, \orgaddress{\street{Colinton Road}, \city{Edinburgh}, \postcode{EH10 5DT}, \country{UK}}}

\abstract{With the increase in deep learning, it becomes increasingly difficult to understand the model in which AI systems can identify objects. Thus, an adversary could aim to modify an image by adding unseen elements, which will confuse the AI in its recognition of an entity. This paper thus investigates the adversarial robustness of LLaVA-1.5-13B and Meta's Llama 3.2 Vision-8B-2. These are tested for untargeted PGD (Projected Gradient Descent) against the visual input modality, and empirically evaluated on the Visual Question Answering (VQA) v2 dataset subset. The results of these adversarial attacks are then quantified using the standard VQA accuracy metric.  This evaluation is then compared with the accuracy degradation (accuracy drop) of LLaVA and Llama 3.2 Vision. A key finding is that Llama 3.2 Vision, despite a lower baseline accuracy in this setup, exhibited a smaller drop in performance under attack compared to LLaVA, particularly at higher perturbation levels. Overall, the findings confirm that the vision modality represents a viable attack vector for degrading the performance of contemporary open-weight VLMs, including Meta's Llama 3.2 Vision. Furthermore, they highlight that adversarial robustness does not necessarily correlate directly with standard benchmark performance and may be influenced by underlying architectural and training factors. }


\keywords{Llama 3.2 Vision, Multimodal Foundation Models, Projected Gradient Descent, Vision-Language Model Safety, Adversarial Examples}

\maketitle

\section{Introduction}

Recent advances in artificial intelligence have led to the development of \emph{foundation models} – large-scale models pre-trained on vast, diverse datasets, exhibiting remarkable capabilities across a range of downstream tasks \citep{bommasaniOpportunitiesRisksFoundation2021}. Prominent examples include Large Language Models (LLMs), which process text, and more recently, Vision-Language Models (VLMs), which integrate visual understanding alongside language capabilities.

The proliferation of these models, particularly powerful open-source variants like Meta's Llama series \citep{touvronLLaMAOpenEfficient2023, touvronLlama2Open2023, grattafioriLlama3Herd2024}, has democratised access but simultaneously introduced significant security considerations \citep{kapoorSocietalImpactOpen2024}. While the security of foundation models is an active research area, much focus has centred on the textual vulnerabilities of LLMs, such as prompt injection or jailbreaking. The integration of vision in VLMs, however, creates an expanded attack surface, introducing vulnerabilities specific to the visual modality that remain comparatively underexplored.

Specifically, VLMs are susceptible to \emph{adversarial image attacks} – carefully crafted, often imperceptible, perturbations applied to input images that can deceive the model and cause erroneous or unintended textual outputs. Understanding this vulnerability is critical, as these multimodal models are increasingly integrated into diverse applications, from content generation to assistive technologies.

This paper addresses the pressing need to assess the adversarial robustness of the \emph{vision component} within state-of-the-art open-weight VLMs. It specifically investigates models related to the influential Llama family, namely LLaVA and Llama 3.2 Vision, and offers one of the first systematic comparisons of visual adversarial robustness in popular open models. This shows that robustness does not always align with standard accuracy metrics. 

\subsection{Aim and objectives}
This paper aims to investigate the security implications arising from the visual modality in these open-source foundation models. The primary focus is on evaluating the adversarial robustness of two prominent VLMs related to Meta's Llama family: LLaVA and Llama 3.2 Vision, using adversarial image attacks. Additionally, the research seeks to analyse how differences in model architecture and training methodologies might influence their robustness against such visual perturbations. The objectives are as follows:

\begin{itemize}
    \item Present the state-of-the-art in the technical evolution of language and computer vision models; integration of multimodality with pre-trained models; and adversarial machine learning attacks \& defences.
    \item Implement LLaVA and Llama 3.2 Vision and test robustness against adversarial examples using Projected Gradient Descent.
    \item Quantify, compare and evaluate the results considering the models' architecture and training.
\end{itemize}

\section{Background}
Recent advancements in artificial intelligence, such as the Transformer \citep{vaswaniAttentionAllYou2017}, have given rise to large language models but also underpin the development of vision-language models. This section lays the groundwork for this evolution, providing the background necessary for language and vision model development, as well as adversarial robustness, to understand the research questions, related work, and methodology of this study.

\subsection{Language Models}

Over the past decade, the field of natural language processing (NLP) has been marked by a shift away from traditional statistical methods towards deep learning models. Prior approaches such as n-gram models \citep{chenEmpiricalStudySmoothing1996}, Hidden Markov Models (HMMs) \citep{rabinerTutorialHiddenMarkov1989}, and phrase-based machine translation \citep{koehnStatisticalPhrasebasedTranslation2003} have now been replaced as state of the art (SOTA) by neural network approaches. 

\subsubsection{Neural Network Approaches to NLP}
Although early neural network research, including backpropagation, dates back to the 1980s \citep{rumelhartLearningRepresentationsBackpropagating1986a}, these models did not achieve widespread use in NLP due to data and hardware limitations. 
Indeed, it was not until 2012 that large-scale deep learning gained mainstream attention in AI research. This breakthrough occurred when AlexNet — a convolutional neural network (CNN) — achieved remarkable results on the 
ImageNet Large-Scale Visual Recognition Challenge (ILSVRC-2012). This annual challenge, considered a benchmark for SOTA computer vision, involved the classification of over a million images across 1,000 object categories. AlexNet achieved a top-5 error rate of only 15.3\%, significantly outperforming the second-best method, which had an error rate of 26.2\% \citep{krizhevskyImageNetClassificationDeep2012}. Although this result occurred in the field of computer vision, it demonstrated the effectiveness of large-scale, GPU-accelerated deep learning and reignited interest in applying neural networks to NLP tasks.

Following the breakthrough, Recurrent Neural Networks (RNNs) or, specifically, a type known as Long Short-Term Memory (LSTM) \citep{hochreiterLongShortTermMemory1997}, originally developed in 1997 to address the vanishing gradient issue, emerged as a promising candidate for applying deep learning to NLP due to their architecture for sequence modelling that captures temporal dependencies. Then, the development of Recurrent Continuous Translation Models \citep{kalchbrennerRecurrentContinuousTranslation2013} led to a significant shift away from phrase-based statistical methods towards neural machine translation (NMT).

The subsequent introduction of the Sequence-to-Sequence (Seq2Seq) framework \citep{sutskeverSequenceSequenceLearning2014a} formalised the encoder-decoder architecture for NMT tasks, significantly improving performance over previous statistical approaches. Nevertheless, the sequential processing characteristic of these models continued to limit their ability to capture long-range dependencies effectively.

\subsubsection{Attention \& The Transformer Architecture}
A significant advancement occurred with the introduction of the attention mechanism in the paper "Neural Machine Translation by Jointly Learning to Align and Translate" \citep{bahdanauNeuralMachineTranslation2016}. Attention allows models to selectively focus on relevant parts of the input sequence when generating output, overcoming the limitations of fixed-length context vectors used in earlier Seq2Seq models. This dynamic weighting mechanism proved crucial for handling long sequences and improving translation quality.

Building upon the concept of attention, the Transformer architecture, introduced in \emph{Attention Is All You Need} \citep{vaswaniAttentionAllYou2017}, revolutionised sequence modelling by entirely replacing recurrent and convolutional layers with self-attention mechanisms. The Transformer consists of an encoder and a decoder, each composed of multiple identical layers.

The encoder maps an input sequence of token embeddings \((x_1, ..., x_n)\) to a sequence of continuous representations \(\mathbf{z} = (z_1, ..., z_n)\). Each encoder layer contains two sub-layers: a multi-head self-attention mechanism and a position-wise fully connected feed-forward network. Each sub-layer is wrapped with a residual connection followed by layer normalisation.

The decoder, conditioned on the encoder outputs \(\mathbf{z}\), generates an output sequence \((y_1, ..., y_m)\) autoregressively, one token at a time. Each decoder layer consists of three sub-layers: a masked multi-head self-attention mechanism (to preserve the autoregressive property), a multi-head cross-attention mechanism over the encoder outputs, and a feed-forward network. As with the encoder, each sub-layer is surrounded by residual connections, followed by layer normalisation.

The core component, self-attention, enables the model to dynamically weigh the relevance of different tokens in a sequence when computing contextualised representations. Multi-head attention extends this mechanism by projecting the input into multiple subspaces, allowing the model to jointly attend to information from different representation subspaces at different positions.

\begin{figure}
    \centering
    \includegraphics[width=0.50\textwidth]{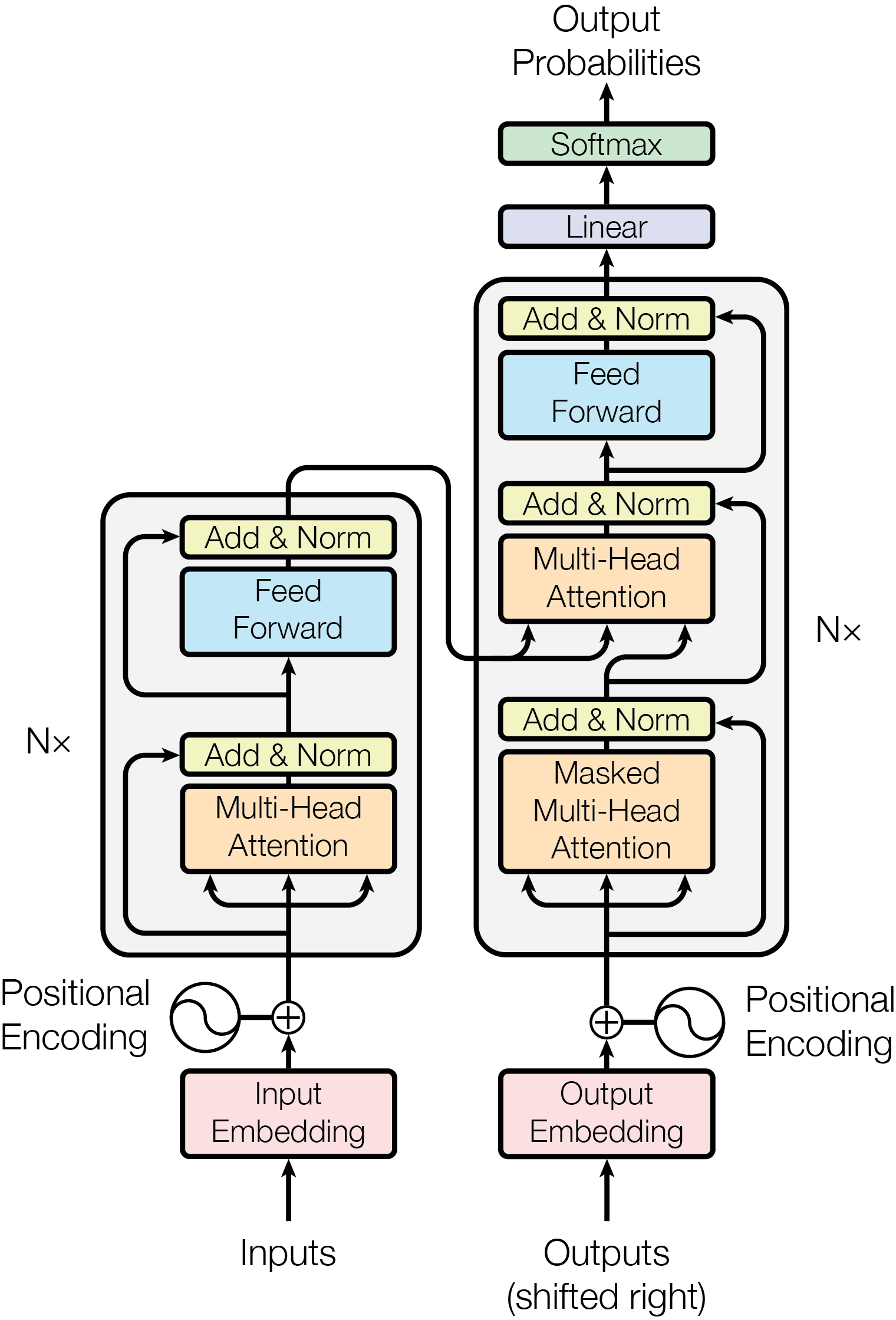}
    \caption{The Transformer architecture illustrates the encoder (left) and decoder (right) stacks with multi-head attention and feed-forward layers. Reproduced from Vaswani et al. (2017) \citep{vaswaniAttentionAllYou2017}.}
    \label{fig:transformer_arch}
\end{figure}

\subsubsection{Foundation Models}
Leveraging the Transformer architecture, the era of foundation models began with OpenAI's Generative Pre-trained Transformer (GPT) \citep{radfordImprovingLanguageUnderstanding2018}. GPT specifically utilised a unidirectional transformer decoder architecture that processed text from left to right. It demonstrated that unsupervised pre-training on vast text corpora followed by supervised fine-tuning could dramatically improve performance across diverse NLP tasks. This model established the present paradigm that dominates the field, allowing models to acquire general language understanding from unlabelled data before adapting to specific downstream applications. With 117 million parameters, GPT achieved state-of-the-art results on numerous language understanding benchmarks.

This pre-training approach was significantly scaled in subsequent models like GPT-2 \citep{radfordLanguageModelsAre2019} and GPT-3 \citep{brownLanguageModelsAre2020}—the latter incorporating approximately 100 times the parameters of GPT-2 (175 billion).

This trend was formalised by the "Scaling Laws for Neural Language Models" paper \citep{kaplanScalingLawsNeural2020}, which showed that performance consistently improves with model size and data volume, leading to the pursuit of ever-larger models.

However, simply scaling models introduced challenges related to safety, alignment with human values, and instruction following. To address this, techniques like Reinforcement Learning from Human Feedback (RLHF) were developed \citep{ouyangTrainingLanguageModels2022}. RLHF fine-tunes models based on human preferences, significantly improving their ability to generate helpful, honest, and harmless responses, becoming a standard practice for state-of-the-art models.

As the rapid pace of development has continued, fierce competition along with safety concerns among leading AI companies has led to the current landscape where powerful closed-source models dominate top benchmarks \citep{wangMMLUProMoreRobust2024, chiangChatbotArenaOpen2024, whiteLiveBenchChallengingContaminationFree2024}. Examples include OpenAI's GPT-4 series \citep{openaiGPT4TechnicalReport2024}, Google's Gemini family, and Anthropic's Claude 3 models. These models exhibit advanced reasoning, multimodal understanding, and near-human performance on various complex tasks. However, their closed nature limits transparency and broad research access.

In parallel, the development of open-weight foundation models has gained significant traction, aiming to democratize access to state-of-the-art AI. Notable examples include Meta's Llama series \citep{touvronLLaMAOpenEfficient2023, touvronLlama2Open2023, grattafioriLlama3Herd2024} and DeepSeek's R1 \citep{deepseek-aiDeepSeekR1IncentivizingReasoning2025}. These models offer competitive performance while fostering wider research, innovation, reproducibility, and community scrutiny.

\subsection{Vision-Language Models}

Vision-language models have developed significantly in the last few years, mirroring the evolution of language models in NLP.

\subsubsection{Early Multimodal Learning}
The integration of vision and language has a long history, with early attempts driven by the need for effective content-based image retrieval (CBIR), where users could search for images using natural language queries. 

Early approaches associated images with textual descriptors by relying on handcrafted features and simple linguistic cues. These methods typically divided images into regions and attempted to map these segments to corresponding words—often nouns and adjectives—to facilitate retrieval tasks \citep{moriImagetowordTransformationBased1999}. While effective in limited domains, such approaches were constrained by their inability to capture the high-level semantics in visual and textual modalities.

The deep learning revolution started by AlexNet \citep{krizhevskyImageNetClassificationDeep2012} in 2012 also had a profound impact on CBIR. The automated extraction of features by Convolutional Neural Networks (CNNs) enabled models to learn complex visual representations directly from raw pixel data and eliminated the need for manual feature engineering. 

A significant step was the introduction of DeViSE (Deep Visual-Semantic Embedding) \citep{fromeDeViSEDeepVisualSemantic2013}, a model designed to align visual features with word embeddings within a shared semantic space. DeViSE employed a CNN to extract image features and a separate neural network for word embeddings, projecting both into a common vector space. The training utilised a contrastive loss function, which encouraged the alignment of related image-text pairs while separating unrelated ones. This enabled zero-shot learning capabilities, allowing the model to recognise object categories not seen during training by leveraging learned semantic relationships. The success of DeViSE highlighted the potential of deep learning to bridge vision and language, influencing subsequent multimodal model development.

As with NLP, a further leap in integrating vision and language was achieved by incorporating attention mechanisms. The paper Show, Attend and Tell \citep{xuShowAttendTell2015} introduced soft and hard attention strategies, enabling models to dynamically focus on regions within images during the caption generation process. This method significantly improved the descriptive accuracy and interpretability of the resulting captions by explicitly aligning image regions with segments of generated text. 

The paper, Deep visual-semantic Alignments for generating image descriptions \citep{karpathyDeepVisualsemanticAlignments2015}, introduced a multimodal recurrent neural network architecture that established strong alignments between specific image regions and corresponding sentence fragments. This approach combined a CNN for visual feature extraction with an RNN for text generation but crucially incorporated a novel alignment objective that matched specific regions in images with words in captions through a common embedding space. By decomposing both images and sentences into more granular components (regions and words) and explicitly learning the correspondence between them, the model achieved more descriptive and accurate caption generation. This alignment approach established a fundamental technique for grounding language in visual content that influenced subsequent work in vision-language modelling.

Following the success of the Transformer architecture in NLP, researchers recognised the potential to extend these models to vision-language tasks. ViLBERT (Vision-and-Language BERT) \citep{luViLBERTPretrainingTaskAgnostic2019} pioneered this extension by building upon the BERT (Bidirectional Encoder Representations from Transformers), the first bidirectional Transformer model \citep{devlinBERTPretrainingDeep2019}. 

ViLBERT introduced a two-stream architecture that allowed for the simultaneous processing of visual and textual information, enabling the model to learn joint representations of both modalities. This was achieved by pre-training on large-scale image-text pairs, allowing the model to learn rich contextual embeddings for both images and text. The model's architecture consisted of two parallel streams: one for visual features extracted from images using CNNs and another for text tokens. The model employed co-attentional transformer layers to facilitate cross-modal interactions, allowing it to learn joint representations that effectively captured the relationships between visual and textual modalities. It demonstrated state-of-the-art performance across multiple vision-language tasks, including visual question answering and visual commonsense reasoning.

Following ViLBERT's dual-stream approach, architectural innovation led to more efficient designs. VisualBERT \citep{liVisualBERTSimplePerformant2019} introduced a unified single-stream architecture that processed visual and textual data within a single Transformer encoder. Rather than maintaining separate streams, it directly combined region-based visual features (extracted via Faster R-CNN) with word embeddings, allowing self-attention to model cross-modal relationships without explicit interaction layers.

UNITER \citep{chenUNITERUNiversalImageTExt2020} further advanced the single-stream approach through four pre-training tasks: masked language modelling, masked region modelling, image-text matching, and optimal transport-based word-region alignment. These objectives enhanced the model's ability to learn fine-grained cross-modal relationships. Experimental results showed that these single-stream architectures consistently outperformed dual-stream counterparts like ViLBERT and LXMERT while using fewer parameters, validating the benefits of unrestricted attention flows between modalities. This architectural shift established a new standard for vision-language integration, particularly for tasks requiring detailed visual reasoning like VQA and image captioning.

Despite these advancements, a common bottleneck persisted: the reliance on CNN-based visual feature extraction. CNNs, while powerful, imposed significant computational overhead and limited scalability when integrated into larger multimodal frameworks. This limitation became increasingly apparent as researchers sought to train on larger datasets, with computational costs becoming a major constraint on further progress.

\subsubsection{Scaling Vision-Language Learning with Transformers}
This limitation was addressed by the Vision Transformer (ViT) introduced in \citep{dosovitskiyImageWorth16x162020}. By partitioning images into fixed-size patches and processing them using self-attention mechanisms, ViT eliminated the need for convolutional operations, thereby aligning the processing paradigms of vision and language under a single Transformer-based framework. The authors demonstrated that when trained on sufficient data, this approach could outperform CNNs while offering better computational scaling properties.

Building upon the architectural innovations of ViT, contrastive learning emerged as a powerful method for training robust vision-language models. CLIP (Contrastive Language-Image Pre-training) \citep{radfordLearningTransferableVisual2021} paired a vision transformer with a GPT-based text encoder in a dual-encoder architecture. By leveraging contrastive objectives trained on 400 million image-text pairs gathered from the internet, CLIP achieved remarkable zero-shot generalisation across a diverse range of tasks. This approach focused on embedding-based retrieval rather than text generation, learning a shared representation space where related image-text pairs were positioned closer together than unrelated ones.

The effectiveness of scale in vision-language learning was further demonstrated by ALIGN \citep{jiaScalingVisualVisionLanguage2021}. Unlike CLIP, ALIGN employed a CNN for visual processing paired with a BERT-based text encoder but similarly used a dual-encoder contrastive approach. Its key innovation was training on an even larger dataset of 1.8 billion image-text pairs with significantly noisier annotations. This work empirically validated that model performance continued to improve with data scale even when using noisier training signals, establishing that data quantity could effectively compensate for reduced annotation quality in multimodal representation learning.

While CLIP and ALIGN demonstrated powerful visual-semantic understanding, they remained limited to retrieval and classification tasks and lacked generative capabilities. SimVLM \citep{wangSimVLMSimpleVisual2022} addressed this limitation by introducing the first unified encoder-decoder visual language model capable of open-ended text generation from visual inputs. SimVLM employed a novel prefix language modelling objective that allowed for more efficient training with weakly supervised image-text pairs. This closed-source model represented a significant milestone in vision-language integration, enabling tasks like image captioning and visual question answering with a single generative framework rather than requiring task-specific fine-tuning as earlier models did.

Building upon the generative capabilities of SimVLM, BLIP (Bootstrapping Language-Image Pre-training) \citep{liBLIPBootstrappingLanguageImage2022} further advanced open-source vision-language modelling. BLIP introduced a novel architecture combining a Vit encoder, a BERT encoder, and a causal language model decoder in a unified framework. It employed a unique bootstrapping approach with two key innovations: a captioning-and-filtering pipeline to generate synthetic captions for web images and a multimodal mixture of encoder-decoder pre-training objectives. This architecture enabled strong performance across image-text retrieval, visual question answering, and image captioning, making it the first open-source model to excel at both discriminative and generative vision-language tasks.

\subsubsection{Multimodal Foundation Models}
While previous vision-language models demonstrated impressive capabilities in task-specific scenarios, they lacked the flexibility to engage in open-ended, multi-turn conversations grounded in a visual context. Flamingo \citep{alayracFlamingoVisualLanguage2022} addressed this limitation by pioneering a fully conversational vision-language model. By integrating a frozen large language model (Chinchilla) with visual inputs through learnable multimodal adapters, Flamingo enabled seamless interactions between text and images within the same conversation flow. The model's perceiver resampler architecture efficiently transformed visual features from a frozen vision encoder into a format compatible with the language model's token representations. Despite being closed-source, Flamingo demonstrated remarkably few-shot learning capabilities across diverse vision-language tasks, setting a new paradigm for multimodal foundation models.

While Flamingo achieved impressive results, its approach remained closed-source. BLIP-2 \citep{liBLIP2BootstrappingLanguageImage2023} introduced an open-source alternative through its innovative Q-Former architecture. This bootstrapping approach connected a frozen ViT encoder to a frozen large language model via a lightweight query transformer module that served as a multimodal adapter. The Q-Former took as input a set of learnable query tokens along with key-value pairs from the vision encoder's embeddings, transforming them into query-aligned image representations compatible with the language model's input space. This design eliminated the need for end-to-end training of massive multimodal models while still achieving state-of-the-art performance on various vision-language tasks, including visual question answering, where it outperformed Flamingo on the VQAv2 benchmark.
Building upon BLIP-2's architecture, LLaVA (Large Language and Vision Assistant) \citep{liuVisualInstructionTuning2023} marked a significant milestone as the first multimodal instruction-tuned model for vision-language tasks. LLaVA's architecture connected a frozen CLIP Vision Encoder (ViT-L/14) with Vicuna—a fine-tuned version of Meta's Llama language model—through a trainable projection layer that mapped visual features into the LLM's word embedding space. This simple yet effective design enabled end-to-end fine-tuning for multimodal instruction-following capabilities.

LLaVA employed a two-stage instruction tuning approach: first, a feature alignment pre-training phase using 595K image-text pairs from CC3M to establish visual-language connections, followed by instruction fine-tuning on 158K GPT-4-generated multimodal instruction-following samples. These samples included conversational QA (58K samples), detailed descriptions (23K samples), and complex reasoning (77K samples). This comprehensive training enabled LLaVA to achieve 85.1\% of GPT-4's performance on multimodal instruction-following tasks while outperforming contemporaries like OpenFlamingo and BLIP-2 in instruction adherence and reasoning. When combined with GPT-4 as an ensemble, LLaVA achieved state-of-the-art performance in Science QA (92.53\% 

A significant evolution in vision-language models arrived with OpenAI's GPT-4V \citep{GPT4VisionSystemCard2024}, also known as GPT-4 Vision. While its architecture remains a closed source, it demonstrates impressive multimodal performance.

The model has shown human-competitive performance on visual benchmarks while also demonstrating emergent capabilities like optical character recognition, multi-image reasoning, and diagram interpretation without explicit training for these tasks. Despite its impressive capabilities, GPT-4V's closed-source nature limits academic scrutiny and understanding of its internal mechanisms, creating challenges for researchers seeking to build upon its innovations.

LLaMA 3.2 Vision \citep{grattafioriLlama3Herd2024} represents Meta's entry into multimodal foundation models with an open-weights approach, in contrast to closed systems like GPT-4V. Rather than developing a native multimodal architecture from scratch, Meta implemented an adapter-based approach that preserves the text capabilities of the pre-trained Llama 3.1 language models.

The architecture consists of three main components: (1) an image encoder based on Vision Transformer (ViT-H/14) with 850M parameters (compared to LLaVA's ViT-L/14), (2) an image adapter with cross-attention layers that integrate visual information into the language model, and (3) a video adapter for temporal reasoning. The cross-attention layers, which add approximately 100B parameters for the Llama 3.1 405B model, are strategically inserted after every fourth self-attention layer in the core language model. This represents a significant architectural difference from LLaVA, which used a simple projection layer to map visual features into the LLM's embedding space.

While LLaVA was trained on 595K image-text pairs from CC3M and 158K GPT-4-generated instruction samples, Meta's training pipeline for Llama 3.2 Vision is substantially more extensive, using approximately 6B image-text pairs for initial pre-training, followed by 500M higher-quality images for adapter annealing. During adapter training, they update the parameters of the image encoder but intentionally preserve the language model parameters, ensuring that the text-only capabilities remain intact. This approach creates a drop-in replacement for Llama 3.1 models that can process both visual and textual inputs while maintaining strong language capabilities.

The post-training phase follows a similar recipe to the text-only models, with several rounds of alignment, including supervised fine-tuning, rejection sampling, and direct preference optimisation, along with safety mitigations to produce a model that balances helpfulness with responsible behaviour.

\subsection{Adversarial Robustness}
Adversarial robustness refers to a model's ability to maintain performance and security when confronted with inputs specifically crafted to deceive it. This property is critical as deep learning systems are increasingly deployed in safety-sensitive domains like autonomous driving, medical diagnostics, and content moderation. For vision-language models (VLMs), robustness is particularly crucial. These models often operate in user-facing applications where vulnerabilities in either modality could lead to significant security risks or unreliable behaviour.

The vulnerability of neural networks to adversarial examples was first systematically demonstrated in the seminal paper "Intriguing Properties of Neural Networks" \citep{szegedyIntriguingPropertiesNeural2014}. The authors discovered that imperceptible, non-random perturbations applied to input images could cause state-of-the-art neural networks to misclassify them with high confidence. These "adversarial examples" were generated by optimising the input image to maximise the model's prediction error, revealing a fundamental weakness. This finding presented a paradox: how could models that generalise well on standard test data be so easily fooled by inputs indistinguishable to humans? 

Subsequent research sought to understand the root causes of this vulnerability. Nguyen et al. \citep{nguyenDeepNeuralNetworks2015} further highlighted the gap between human and machine perception by showing that deep neural networks could confidently classify completely unrecognisable, abstract patterns as familiar objects. This suggested that networks might rely on superficial statistical correlations rather than robust, human-like semantic understanding.

Goodfellow et al. \citep{goodfellowExplainingHarnessingAdversarial2015} proposed the "linear explanation," arguing that the high dimensionality of input spaces makes even models with predominantly linear behaviour susceptible. Small perturbations across many dimensions can accumulate to cause large changes in the output, even if the model behaves linearly locally.

These discoveries have spurred the growth of Adversarial Machine Learning (AML) as a field dedicated to studying attacks on machine learning systems and developing defences. 

A particularly challenging property, identified early on in one study \citep{szegedyIntriguingPropertiesNeural2014}, is that adversarial examples crafted to fool one model often successfully deceive other models. This is often even for those with different architectures or trained on different data subsets. This cross-model vulnerability suggests that adversarial examples exploit fundamental characteristics of the data distribution or the learning process itself rather than just overfitting to specific model parameters.

The existence of such adversarial examples has profound implications for real-world systems beyond the theory of visual perturbations, causing incorrect image interpretations. Understanding these vulnerabilities is the first step towards building more secure and reliable AI systems, and this necessitates a structured approach to classifying and analysing these threats.

\subsubsection{Taxonomy of Adversarial Machine Learning}

One of the earliest structured taxonomies to categorise attack methodologies systematically \citep{pitropakisTaxonomySurveyAttacks2019} defined adversarial attacks based on two primary phases: preparation and manifestation. Preparation includes access to the model, leading to the distinction between White-box attacks, where the attacker has complete knowledge of the model's architecture, weights, and gradients, and Black-box attacks, where no internal model details are accessible. The manifestation categorises adversarial attacks by their specificity (targeted vs untargeted), mode (colluding vs non-colluding), and type (poisoning/training time attacks vs evasion/run time attacks). This taxonomy established a foundation for categorising threats against machine learning systems.

More recently, the National Institute of Standards and Technology (NIST) introduced a comprehensive classification formalising an industry-wide standard for evaluating both attacks and defences \citep{vassilevAdversarialMachineLearning2024}. This taxonomy provides a structured approach to understanding the entire adversarial machine learning landscape, covering attack vectors, defence mechanisms, and evaluation criteria across different domains and modalities.

\subsubsection{Formalisation of Threat Models}
A formalised approach to threat models is essential for the rigorous evaluation of adversarial robustness. Carlini and Wagner \citep{carliniEvaluatingAdversarialRobustness2019} provided a landmark contribution by outlining comprehensive best practices and establishing a framework for evaluating defences. They formalised the adversarial robustness problem as finding the minimal perturbation $\delta$ that causes misclassification:

\begin{align}
\text{minimise } &\|\delta\|_p\\
\text{such that } &f(x + \delta) \neq f(x)\\
&x + \delta \in [0,1]^n \label{eq:adv_robustness}
\end{align}

Where $f$ is the model, $x$ is the input, $\delta$ is the perturbation, $\|\delta\|_p$ represents the $L_p$ norm measuring perturbation magnitude, and the final constraint ensures the perturbed input remains valid (e.g., pixel values within range).

\subsubsection{Defence Evaluation Principles}

Their work established three core principles for rigorous evaluation:

\begin{itemize}
    \item \textbf{Threat Model Specification}: Clearly defining adversarial goals (e.g., misclassification), knowledge (white-box, black-box, grey-box), and capabilities (e.g., perturbation budget specified by $L_p$ norm).
    \item \textbf{Adaptive Adversary Attacks}: Testing against optimised attacks specifically designed to overcome the proposed defence, rather than relying solely on standard, pre-existing attacks. An adaptive adversary modifies attack strategies based on knowledge of the defence mechanism.
    \item \textbf{White-Box vs Black-Box Testing}: Ensuring defences are robust under both scenarios. Defences that only work in black-box settings might rely on "security through obscurity" (e.g., gradient masking) rather than true robustness.
\end{itemize}

\subsubsection{Common Evaluation Pitfalls}

Carlini and Wagner identified critical pitfalls that can lead to overestimating a defence's effectiveness:

\begin{itemize}
    \item \textbf{Gradient Masking/Obfuscation}: Some defences inadvertently (or deliberately) make it harder to find useful gradients, hindering gradient-based attacks. This creates an illusion of security that often fails against gradient-free methods, transfer attacks, or adaptive white-box attacks.
    \item \textbf{Weak Attack Configuration}: Using suboptimal attack parameters (e.g., insufficient iterations, inappropriate step sizes, or incorrect loss functions for the attack goal) can make a defence appear stronger than it is.
    \item \textbf{Inadequate Testing Scope}: Failing to test against a diverse range of attack types, particularly adaptive attacks tailored to the defence mechanism.
\end{itemize}

\section{Literature Review}

Building upon the foundational concepts introduced in the previous section, this section examines specific research directly relevant to adversarial examples against vision components in multimodal systems, detailing attack methodologies that form the basis for the experimental approach.

\subsection{Types of Adversarial Examples in Computer Vision}
Adversarial examples in computer vision can be generated through various techniques, each with different computational requirements and effectiveness.

\subsubsection{Fast Gradient Sign Method (FGSM)}
\begin{figure}
    \centering
    \subfloat[Original Image: "Panda" (57.7\% confidence)]{%
        \includegraphics[width=0.3\textwidth]{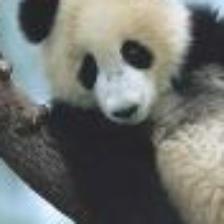}%
        \label{fig:panda_original}%
    }\hfill
    \subfloat[Adversarial Noise ($\epsilon = 0.007$)]{%
        \includegraphics[width=0.3\textwidth]{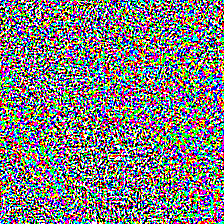}%
        \label{fig:panda_noise}%
    }\hfill
    \subfloat[Adversarial Image: "Gibbon" (99.3\% confidence)]{%
        \includegraphics[width=0.3\textwidth]{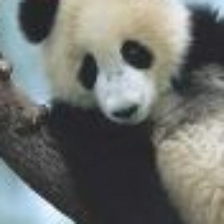}%
        \label{fig:panda_adversarial}%
    }
    \caption{Demonstration of adversarial perturbation using FGSM. An imperceptible perturbation $\eta = \epsilon \cdot \text{sign}(\nabla_x J(\theta, x, y))$ is added to the original image (a), resulting in an adversarial image (c) that causes misclassification with high confidence. The noise (b) is scaled for visibility. Reproduced from Goodfellow et al. (2015) \citep{goodfellowExplainingHarnessingAdversarial2015}.}
    \label{fig:fgsm_3panel}
\end{figure}

One of the earliest and most influential methods for generating adversarial examples is the Fast Gradient Sign Method (FGSM), introduced by Goodfellow et al. \citep{goodfellowExplainingHarnessingAdversarial2015}. FGSM operates on the principle that neural networks, particularly in high-dimensional spaces, behave in a surprisingly linear fashion. This linearity allows for the efficient generation of adversarial perturbations by taking a single step in the direction that maximises the model's loss. The perturbation \(\eta\) is calculated as:
\[
\eta = \epsilon \cdot \text{sign}\!\bigl(\nabla_x J(\theta, x, y_{\text{true}})\bigr)
\]
where \(x\) is the original input image, \(y_{\text{true}}\) is the true label, \(\theta\) represents the model parameters, \(J\) is the loss function (e.g., cross-entropy), \(\nabla_x J\) is the gradient of the loss with respect to the input image \(x\), \(\text{sign}(\cdot)\) is the sign function, and \(\epsilon\) is a small scalar controlling the magnitude of the perturbation (often constrained by the $L_\infty$ norm). The adversarial example \(x_{\text{adv}}\) is then created by adding this perturbation to the original image:
\[
x_{\text{adv}} = x + \eta
\]
Despite its simplicity and computational efficiency (requiring only one gradient computation), FGSM proved remarkably effective at fooling networks, demonstrating that adversarial vulnerability was a fundamental property linked to linearity in high dimensions \citep{goodfellowExplainingHarnessingAdversarial2015}. Figure \ref{fig:fgsm_3panel} illustrates how a small, visually imperceptible FGSM perturbation can cause a model to misclassify an image with high confidence.

\subsubsection{Carlini \& Wagner Attacks}
While FGSM provides a fast way to generate adversarial examples, Carlini \& Wagner \citep{carliniEvaluatingRobustnessNeural2017} treat the generation process as an optimization problem, aiming to find the minimal perturbation required to cause misclassification according to a specific distance metric (e.g., $L_0$, $L_2$, $L_\infty$).

The author's family of attacks have come to be known as Carlini \& Wagner (CW) attacks \citep{carliniEvaluatingRobustnessNeural2017} and are designed to be particularly effective against defensive techniques like defensive distillation, which FGSM might fail to overcome. The CW attacks, especially the $L_2$ version, formulate the problem as minimising the perturbation norm $\|\delta\|_2$ subject to the constraint that the perturbed input $x+\delta$ is misclassified. This is often achieved by minimising a loss function that encourages misclassification while simultaneously penalising the perturbation size:
\[
\min_{\delta} \|\delta\|_2^2 + c \cdot \mathcal{L}(x+\delta)
\]
where $\mathcal{L}$ is a loss function designed such that $\mathcal{L}(x') \le 0$ if and only if the model classifies $x'$ incorrectly (for targeted attacks, incorrectly as the target class), and $c$ is a constant balancing the two terms. CW attacks are generally more powerful but computationally more expensive than FGSM, often serving as a stronger benchmark for evaluating defences \citep{carliniEvaluatingRobustnessNeural2017}.

\subsubsection{Projected Gradient Descent (PGD)}
Projected Gradient Descent (PGD) \citep{madryDeepLearningModels2019} is arguably the most widely used and powerful first-order adversarial attack, building upon the ideas of FGSM but employing an iterative approach. Instead of taking one large step, PGD takes multiple smaller steps in the direction of the gradient sign, projecting the result back onto the allowed perturbation region (e.g., an $L_\infty$ ball of radius $\epsilon$ around the original input $x$) after each step. The update rule for each iteration $t$ is:
\[
x^{t+1} = \Pi_{\mathcal{B}_\epsilon(x)} \left( x^t + \alpha \cdot \text{sign}\!\bigl(\nabla_{x^t} J(\theta, x^t, y_{\text{true}})\bigr) \right)
\]
Here, $x^0$ is typically initialised randomly within the perturbation ball $\mathcal{B}_\epsilon(x)$, $\alpha$ is the step size (usually smaller than $\epsilon$), and $\Pi_{\mathcal{B}_\epsilon(x)}$ denotes the projection operator that ensures the perturbed image $x^{t+1}$ remains within the $\epsilon$-ball around the original image $x$.

By using multiple iterations and random starts, PGD is much more effective at finding adversarial examples than FGSM, especially for models trained to be robust. Madry et al. \citep{madryDeepLearningModels2019} framed adversarial robustness within a min-max optimisation perspective, arguing that PGD acts as a universal first-order adversary capable of finding the worst-case perturbation within the specified threat model. Consequently, PGD has become the standard benchmark attack for evaluating adversarial defences and is commonly used in adversarial training regimes to improve model robustness.

\subsection{Multimodal Foundation Model Robustness}

\subsubsection{Expanded Attack Surface}
Multimodal Foundation Models inherently present a more complex security landscape than their unimodal counterparts. By processing both visual and textual inputs simultaneously, these models effectively incorporate additional attack vectors for potential exploitation, analogous to adding more doors or windows to a house, resulting in additional points of entry. This expanded attack surface creates unique vulnerabilities at the intersection of modalities, where adversarial manipulations in one domain can propagate to and influence the other and potentially bypass unimodal specific defences.

For example, in one paper \citep{liuMMSafetyBenchBenchmarkSafety2025}, the authors demonstrated that query-relevant images could bypass text-based safety measures and produce unsafe output in multimodal models without even requiring adversarial perturbations. Using three image generation methods, they found an average increase in Attack Success Rate (ASR), exceeding 30\% in LLaVA-1.5-7B, and that text-only alignment proves insufficient for securing these systems.

Furthermore, despite the growing prevalence of multimodal systems, research on foundation model robustness has primarily focused on text-based attacks, such as prompt injection and jailbreaking \citep{shinAutoPromptElicitingKnowledge2020,wallaceUniversalAdversarialTriggers2021, weiJailbrokenHowDoes2023,zouUniversalTransferableAdversarial2023}. This emphasis on textual vulnerabilities has left visual attack vectors comparatively underexplored, creating a potential blind spot in security evaluations. The integration of vision capabilities potentially introduces adversarial examples, as discussed in relation to computer vision systems as an attack vector to foundation models.

\subsubsection{Cross-Modal Adversarial Attacks}
The integration of vision and language creates vulnerabilities where adversarial manipulations in one modality can not only influence but potentially jailbreak the other. The authors of Image Hijacks: Adversarial Images Can Control Generative Models at Runtime \citep{lukebaileyImageHijacksAdversarial2023} established that carefully crafted pixel perturbations could effectively hijack the behaviour of Vision-Language Models (VLMs). The authors conducted white-box attacks against LLaVA (specifically, the LLaMA-2-13B-Chat language model combined with a CLIP ViT-L/14 vision encoder), employing their "Behaviour Matching" algorithm—a targeted approach based on Projected Gradient Descent (PGD)—to generate adversarial images. These images contained subtle perturbations, constrained within $L_{\infty}$ norms ranging from 1/255 to 64/255, yet were sufficient to induce specific, targeted behaviours in the model. Their experiments demonstrated high success rates (exceeding 80\%) across four distinct attack types: forcing the VLM to output exact predefined adversarial messages (Specific String Attack), compelling the model to expose private information within API calls (Context Leak Attack), bypassing RLHF-based safety guardrails to generate harmful content (Jailbreak Attack), and embedding factual distortions into images to modify the model's perceived world knowledge (Disinformation Attack). This work highlighted the significant risk posed by adversarial images, showing they could reliably control the textual output of generative VLMs at runtime.

Concurrently with this paper, a study \citep{schlarmannAdversarialRobustnessMultiModal2023} further investigated similar vulnerabilities through an evaluation of the adversarial robustness of OpenFlamingo, an open-source implementation of the Flamingo architecture. Their work demonstrated that even imperceptible adversarial perturbations (as small as $\epsilon = 1/255$) could fully manipulate generated captions. They employed both targeted attacks, designed to force specific adversary-defined outputs like fake news, and untargeted attacks aimed at degrading output quality. Achieving high success rates in image captioning and Visual Question Answering (VQA) tasks, their findings underscore the significant real-world security risks posed by such attacks, including misinformation propagation and user manipulation.

The paper Jailbreak in Pieces: Compositional Adversarial Attacks on Multimodal Language models \citep{shayeganiJailbreakPiecesCompositional2023} introduced "compositional adversarial attacks" that exploit cross-modal alignment vulnerabilities in Vision-Language Models (VLMS). Their approach paired benign textual prompts with adversarial images, effectively bypassing text-only safety measures without requiring white-box access to the language model component. By targeting the embedding space through four distinct strategies (textual triggers, OCR textual triggers, visual triggers, and combined OCR+visual triggers), they achieved attack success rates of up to 91\% on LLaVA. Notably, these attacks demonstrated that the same adversarial image could jailbreak multiple textual prompts, and once a model was compromised, subsequent prompts continued generating harmful outputs—a phenomenon termed "context contamination.

Expanding investigation to commercial systems, researchers evaluated the black-box adversarial robustness of Google's Bard, demonstrating that adversarial image perturbations could manipulate model responses with a 22\% attack success rate on image descriptions \citep{dongHowRobustGoogles2023}. Their study revealed vulnerabilities in Bard's face detection and toxicity detection defences and demonstrated attack transferability across other commercial systems, including GPT-4V (45\% success), Bing Chat (26\%), and ERNIE Bot (86\%). This work highlighted that even sophisticated closed-source models remain vulnerable to adversarial manipulation through their visual components despite employing robust text-based safety filters.

Carlini et al. \citep{carliniAreAlignedNeural2023} investigated whether alignment techniques like RLHF confer robustness specifically against text-based adversarial attacks. They demonstrated that current text-based attacks often fail against aligned language models, not due to inherent robustness from alignment but because the existing NLP attack methods themselves lack sufficient power. In contrast, the research highlighted that multimodal vision-language models could still be controlled via subtle image perturbations, even when aligned. These findings question whether alignment truly hardens models against sophisticated textual manipulation or merely addresses superficial behaviours, suggesting stronger text-based attacks might eventually bypass alignment in text-only models. Understanding precisely where this increased vulnerability from multimodality arises is an important area for future work, as it is likely that future models incorporating additional modalities, such as audio, will introduce new vulnerabilities and expand the attack surface \citep{carliniAreAlignedNeural2023}.

A different study \citep{bhagwatkarImprovingAdversarialRobustness2024} investigated architectural and prompt design factors affecting adversarial robustness in vision-language models. Applying PGD attacks against several open foundation models, including LLaVA but not Llama 3.2 vision, they made two key discoveries: first, contrary to intuition, neither increasing the input resolution of vision encoders nor scaling up the language model component enhanced robustness against adversarial examples. Second, they found that certain prompt design choices, particularly the inclusion of task-specific instructions and contextual information, could marginally improve robustness in some settings.

In response to concerns about visual input safety, Meta released their own multimodal content moderation model, Llama 3 Guard Vision \citep{chiLlamaGuard32024}, designed to classify and prevent harmful interactions with their Llama 3.2 vision model, similar to how their LLM-based model, Llama Guard \citep{inanLlamaGuardLLMbased2023} safeguards their conversational models. Llama Guad Vision was fine-tuned on 22,500 multimodal prompts and 40,034 response examples, outperforming GPT-4o in safety classification with higher F1 scores and lower false positive rates. The authors evaluated Llama guard vision's robustness in this paper \citep{chiLlamaGuard32024} against adversarial attacks, showing that PGD image perturbations $(\varepsilon=8/255)$ successfully increased harmful content misclassification rates from 21\% to 70\%. The paper concludes that Meta recommends adversarial training to improve robustness before deploying and limiting model access to images in safety-critical applications - interesting for a pre-trained model designed to be a drop-in security enhancement for Llama 3.2 vision.

\section{Methodology}
This paper investigates the security of vision in state-of-the-art open-source foundation models by evaluating the adversarial robustness of popular open-source models related to Meta's Llama family. Specifically, we employ adversarial image attacks against LLaVa and Llama 3.2 Vision. Additionally, the research seeks to explore how model architecture and training influence robustness by analysing their relationship with adversarial performance. 

\subsection{Problem setting}
The objectives of the methodology are defined in this section. 
\subsubsection{Formal Definition}
Let $f$ represent a VLM, which takes an image $x \in \mathbb{R}^{H \times W \times C}$ (where H, W, C are height, width, channels) and a text prompt $q$ as input, and generates a textual output sequence $y = f(x, q)$. The goal of an adversarial attack in this context was to find a small perturbation $\delta$, such that when added to the original image $x$, the resulting adversarial image $x_{adv} = x + \delta$ caused the model to produce an undesirable or incorrect output $y_{adv} = f(x_{adv}, q)$, where $y_{adv}$ significantly differed from the expected output $y = f(x, q)$. The magnitude of the perturbation $\delta$ was constrained, typically using an $L_p$ norm, to ensure it remained imperceptible or near-imperceptible to humans.

\subsubsection{Models Under Investigation}
This study focused on two prominent open-weight Foundation Models from the Llama family:
\begin{itemize}
    \item \textbf{LLaVA (Large Language and Vision Assistant) v1.5-13B}: An established VLM that connected a pre-trained CLIP ViT-L/14 visual encoder to a Vicuna-13B language model (a Llama variant) via a simple projection layer. It underwent instruction tuning on multimodal datasets \citep{liuVisualInstructionTuning2023}. 
    \item \textbf{Llama 3.2 Vision-8B-2}: Meta's first open multimodal foundation model, which integrated a ViT-H/14 image encoder with a Llama 3.1 8B language model using a more complex cross-attention adapter mechanism. It was pre-trained and aligned on significantly larger datasets \citep{grattafioriLlama3Herd2024}.
\end{itemize}
These models represented different architectural approaches (simple projection vs. cross-attention adapter) and training scales, allowing for comparative analysis of robustness.

\subsubsection{Notation}
\begin{itemize}
    \item $x$: Original input image.
    \item $q$: Input text prompt.
    \item $y$: Generated text output from the VLM for the original input, $y = f(x, q)$.
    \item $\theta$: Parameters of the VLM $f$.
    \item $\delta$: Adversarial perturbation added to the image.
    \item $x_{adv}$: Adversarial image, $x_{adv} = x + \delta$.
    \item $y_{adv}$: Generated text output for the adversarial input, $y_{adv} = f(x_{adv}, q)$.
    \item $\epsilon$: Maximum perturbation budget, constraining $\|\delta\|_p \le \epsilon$.
    \item $J(\theta, x, q, y_{target})$: Loss function used to guide the attack, measuring the discrepancy between the model's prediction and a target (or lack thereof for untargeted attacks).
\end{itemize}

\subsubsection{Threat Model}
Following the best practices outlined by Carlini and Wagner \citep{carliniEvaluatingAdversarialRobustness2019}, the threat model was defined as follows:
\begin{itemize}
    \item \textbf{Goal}: Untargeted attack. The objective was not to force the VLM to output a specific incorrect string but rather to degrade the quality, relevance, or correctness of the generated text $y_{adv}$ compared to the expected output $y$ for the clean image $x$ and prompt $q$. This was often achieved by maximising the model's internal loss function with respect to the input image.
    \item \textbf{Knowledge}: White-box access. The attacker was assumed to have complete knowledge of the VLM's architecture, parameters ($\theta$), and gradients ($\nabla_x J$). This allowed for the use of powerful gradient-based attack methods like PGD.
    \item \textbf{Capabilities}: The attacker could only modify the input image $x$ by adding a perturbation $\delta$. The text prompt $q$ remained unchanged. The perturbation was constrained by the $L_\infty$ norm, such that $\|\delta\|_\infty \le \epsilon$. This ensured that the changes to individual pixel values were bounded, maintaining the visual similarity between $x$ and $x_{adv}$. The value of $\epsilon$ defined the strength of the attack.
\end{itemize}

\subsubsection{Assumptions}
\begin{itemize}
    \item The attack vector was solely the visual input ($x$). The textual prompt ($q$) was assumed to be benign and fixed for each image-question pair during the attack generation and evaluation.
    \item The Visual Question Answering (VQA) v2 dataset \citep{goyalMakingVQAMatter2017} served as the benchmark task. Performance degradation on this task was considered indicative of reduced model robustness, as VQA required integrated visual and language understanding.
    \item The effectiveness of the attack was measured by the degradation in the model's ability to correctly answer the questions associated with the images after perturbation.
\end{itemize}

\subsection{Method}
This study employed Projected Gradient Descent (PGD) as the primary method for generating adversarial examples targeting the visual input of the selected VLMs.

\subsubsection{Attack Algorithm: Projected Gradient Descent (PGD)}
PGD was chosen due to its established effectiveness as a powerful, first-order white-box adversarial attack \citep{madryDeepLearningModels2019}. As discussed in the Background section, PGD built upon the Fast Gradient Sign Method (FGSM) \citep{goodfellowExplainingHarnessingAdversarial2015} but utilised an iterative approach with smaller steps and projection, which made it significantly more effective at finding adversarial examples, particularly against models potentially incorporating defensive measures. Madry et al. \citep{madryDeepLearningModels2019} demonstrated that PGD served as a universal first-order adversary, capable of finding approximate worst-case perturbations within a given threat model ($L_\infty$ ball in this case). Its widespread adoption as a standard benchmark for evaluating adversarial robustness made it suitable for rigorously assessing and comparing the resilience of LLaVA and Llama 3.2 Vision.

\subsubsection{PGD Implementation for Generative VLMs}
The standard PGD algorithm was adapted here for the context of generative VLMs performing Visual Question Answering (VQA).

\subsubsection{Objective}
Consistent with the untargeted attack goal defined in the Threat Model, the objective was to find a perturbation $\delta$ that, when added to the original image $x$, maximised the model's internal loss function for the given image-prompt pair $(x, q)$. The aim was not to force a specific incorrect output string but to generate an adversarial image $x_{adv} = x + \delta$ that caused the VLM $f$ to produce a degraded, irrelevant, or incorrect textual output $y_{adv} = f(x_{adv}, q)$ compared to the expected output $y = f(x, q)$.

\subsubsection{Loss Function}
The attack leveraged the model's internal loss function, typically related to the negative log-likelihood or cross-entropy of generating the output sequence given the inputs. For an untargeted attack on the image modality, the PGD algorithm aimed to maximise this loss $J(\theta, x_{adv}, q)$ with respect to the input image $x_{adv}$. The gradient $\nabla_{x_{adv}} J$ indicated the direction in the image space that most increased the model's loss (i.e., made the current prediction less likely or increased model uncertainty/error signal).

\subsubsection{Optimisation Process}
The PGD attack iteratively refined the perturbation $\delta$. Starting with an initial perturbed image $x^0$ (often $x$ or $x$ plus small random noise within the $\epsilon$-ball), each iteration updated the image as follows:
\[
x^{t+1} = \Pi_{\mathcal{B}_\epsilon(x)} \left( x^t + \alpha \cdot \text{sign}\!\bigl(\nabla_{x^t} J(\theta, x^t, q)\bigr) \right)
\]
Where:
\begin{itemize}
    \item $x^t$ was the adversarial image at iteration $t$.
    \item $\alpha$ was the step size, determining how large a step was taken in the gradient's sign direction.
    \item $\nabla_{x^t} J(\theta, x^t, q)$ was the gradient of the loss function with respect to the image input $x^t$, computed using the model parameters $\theta$ and the fixed prompt $q$.
    \item $\text{sign}(\cdot)$ extracted the direction of steepest ascent for the loss.
    \item $\Pi_{\mathcal{B}_\epsilon(x)}$ was the projection operator. After each step, it ensured the resulting image $x^{t+1}$ remained within the $L_\infty$ ball of radius $\epsilon$ centred around the original image $x$. This meant for every pixel channel $i$, $|x^{t+1}_i - x_i| \le \epsilon$. It also clipped pixel values to the valid range (e.g., [0, 1] for normalised images).
\end{itemize}
This process was repeated for a predefined number of iterations.

\subsubsection{PGD Hyperparameters}
The effectiveness and characteristics of the PGD attack were controlled by several hyperparameters:
\begin{itemize}
    \item \textbf{Perturbation Budget ($\epsilon$)}: Defined the maximum allowed $L_\infty$ distance between the original image $x$ and the adversarial image $x_{adv}$. A larger $\epsilon$ allowed for stronger attacks but potentially more perceptible perturbations.
    \item \textbf{Step Size ($\alpha$)}: Controlled the magnitude of the update at each iteration. It was typically set to a value smaller than $\epsilon$ (e.g., $\alpha = \epsilon / k$ for some $k$).
    \item \textbf{Number of Iterations}: Determined how many gradient ascent steps were performed. More iterations generally led to stronger attacks but increased computational cost.
\end{itemize}
The specific values chosen for these hyperparameters in the experiments were detailed in the following Experimental Setup section.

\subsection{Experimental setup}

\subsubsection{Models}
The experiments utilised the two open-weight VLMs defined in the Problem Setting section, loaded via the Hugging Face Transformers library with the following identifiers:
\begin{itemize}
    \item \textbf{LLaVA-1.5-13B}, Identifier: llava-hf/llava-1.5-13b-hf.
    \item \textbf{Llama 3.2 Vision-8B-2}, Identifier: meta-llama/Llama-3.2-Vision-8B-2.
\end{itemize}
Both models were loaded using \texttt{float16} precision for computational efficiency, as specified in the project's configuration file (\texttt{config.json}). The choice of these models allowed for a direct comparison between an established VLM architecture (LLaVA) and Meta's newer, adapter-based approach (Llama 3.2 Vision).

\subsubsection{Dataset}

The study was conducted using the Visual Question Answering (VQA) v2 dataset \citep{goyalMakingVQAMatter2017}, specifically the validation split. This large-scale dataset contained open-ended questions about images from the Microsoft COCO collection \citep{linMicrosoftCOCOCommon2015}, which required an understanding of vision, language, and commonsense knowledge to answer. The complete dataset included over 200,000 images from Microsoft COCO, more than 1 million questions, and over 10 million answers across diverse question types requiring reasoning, recognition, and understanding of visual content.

Due to the computational demands of generating adversarial examples for the entire dataset, a subsetting approach was employed, as implemented in \texttt{create\_vqav2\_subset.py}. This script randomly sampled a specified number of image-question-answer triplets from the full validation set, which created reproducible subsets across runs. The size of this subset was controlled by the \texttt{subset\_size} parameter in the \texttt{config.json} file. For the experiments reported in this study, a subset size of 500 samples was used, providing a balance between computational feasibility and statistical significance with margins of error around ±3-4\%. Each sample consisted of an image, a question, and a list of ground-truth answers provided by human annotators.

\subsubsection{Evaluation Metrics}
The primary metric used to evaluate model performance and robustness was the standard VQA accuracy, consistent with the VQA v2 challenge evaluation protocol \citep{goyalMakingVQAMatter2017}. This metric measured the agreement between the model's generated answer and the set of human-provided ground-truth answers for each question.

Calculating the VQA accuracy involved normalising both the predicted answer and the ground-truth answers to account for variations in phrasing, case sensitivity, and punctuation. The evaluation process was implemented in the \texttt{compute\_vqa\_accuracy} function within \texttt{vqa\_utils.py}, which was adapted from the original VQA evaluation code \citep{goyalMakingVQAMatter2017}.

An answer was considered correct if the normalised predicted answer matched at least one of the normalised ground-truth answers exactly or if either the prediction was a substring of a ground truth or vice-versa, allowing for minor variations. An empty prediction was always marked as incorrect.

This VQA accuracy was calculated for each model on both the original (clean) images and the corresponding adversarial images generated using PGD. The difference between the average clean accuracy and the average adversarial accuracy, termed the \emph{accuracy drop}, served as the key indicator of the model's susceptibility to the adversarial attack under a specific perturbation budget ($\epsilon$). A larger accuracy drop indicated lower adversarial robustness. The evaluation process was orchestrated by the \texttt{run\_eval.py} script, which computed and recorded these metrics for each parameter set defined in \texttt{config.json}.

\subsubsection{Implementation details}
The experiments were implemented in Python 3, leveraging several key libraries from the scientific computing and machine learning ecosystem. The core deep learning framework used was PyTorch, along with the Hugging Face \texttt{transformers} library for model loading (\texttt{model\_factory.py}) and processing, and \texttt{accelerate} for efficient hardware utilisation. Image processing relied on the Pillow (PIL) library.

The PGD attack logic was implemented in \texttt{pgd\_utils.py}, following the iterative process described in the Method section. The attack utilised the model's internal loss function during backpropagation to compute gradients with respect to the input image pixels. Mixed-precision training (\texttt{float16}, specified in \texttt{config.json}) and gradient checkpointing (\texttt{model\_factory.py}) were employed to manage the significant memory requirements of the large VLMs.

The main evaluation loop was orchestrated by \texttt{run\_eval.py}. This script handled loading the VQA subset, iterating through samples, generating both clean and adversarial predictions using the \texttt{model\_infer.py} module, computing VQA accuracy via \texttt{vqa\_utils.py}, and saving results. Model-specific prompting and processing logic were handled within \texttt{model\_infer.py} and \texttt{run\_eval.py} to ensure correct input formatting for both LLaVA and Llama 3.2 Vision.

The PGD attack hyperparameters were defined in \texttt{config.json} under \texttt{pgd \_parameters\_sets}. Experiments were conducted across a range of perturbation budgets ($\epsilon$), specifically:
\[
\epsilon \in \{2/255, 4/255, 8/255, 16/255, 128/255, 255/255\} 
\]

The step size ($\alpha$) and number of iterations were adjusted proportionally to $\epsilon$, ranging from $\alpha=0.00196$, iterations=5 for $\epsilon=2/255$ up to $\alpha=0.06274$, iterations=30 for $\epsilon=255/255$, ensuring a consistent attack strength relative to the budget. All experiments were executed on an NVIDIA A100 GPU possessing 80GB of VRAM, as required by the memory demands of the 13B and 8B parameter models under evaluation, particularly during the gradient computations needed for the PGD attack.

\section{Evaluation}
This section focuses on quantifying the impact of Projected Gradient Descent (PGD) adversarial attacks on the performance of LLaVA-1.5-13B and Llama 3.2 Vision-8 B-2 on the Visual Question Answering (VQA) task. The section begins by presenting the baseline performance of each model on the clean VQA v2 subset, followed by their performance under adversarial conditions with varying perturbation budgets ($\epsilon$). A comparative analysis then examines the relative robustness of the two models based on the observed accuracy degradation. Finally, a critical analysis discusses the implications of these findings in the context of the research questions, relates them to existing literature, and acknowledges the limitations of the study.

\subsection{Results}
The adversarial robustness of LLaVA-1.5-13B and Llama 3.2 Vision-8B-2 was evaluated on a subset of 500 samples from the VQA v2 validation dataset. Performance was measured using the standard VQA accuracy metric on both original (clean) images and adversarial images generated via Projected Gradient Descent (PGD) with varying $L_\infty$ perturbation budgets ($\epsilon$). The accuracy drop (Clean Accuracy - Adversarial Accuracy) quantifies the impact of the attack.

On the clean dataset subset, LLaVA-1.5-13B achieved a baseline VQA accuracy of 87.4\%. Llama 3.2 Vision-8B-2 achieved a baseline VQA accuracy of 42.8\% on the same subset. It was noted that the observed baseline accuracy for Llama 3.2 Vision (42.8\%) was substantially lower than the 75.2\% reported by Meta on the full VQA v2 dataset \citep{grattafioriLlama3Herd2024}. In contrast, the observed accuracy for LLaVA (87.4\%) was closer to benchmarks reported for similar models (around 84\% \citep{liuVisualInstructionTuning2023}). This significant discrepancy for Llama 3.2 Vision likely stems primarily from implementation-specific factors, such as deviations from the precise, prompt formatting, image preprocessing, or generation parameters required by the model for optimal VQA performance rather than solely subset sampling effects (which typically yield margins of error around ±3-4\% for this size). Additionally, LLaVA's explicit instruction tuning on QA datasets and its larger 13B language model backbone (compared to Llama 3.2 Vision's 8B) might have contributed to its higher baseline performance in this specific experimental setup. Nevertheless, as the primary goal of this study was to assess the relative change in accuracy under adversarial attack, the observed baseline served as the reference point for evaluating robustness, mitigating the impact of this discrepancy on the comparative findings regarding accuracy drop.

Furthermore, a minor variation was observed in Llama 3.2 Vision's baseline accuracy during the run with the largest perturbation budget $\epsilon = 255/255$. While the baseline was consistently 42.8\% across other runs, it was recorded as 41.6\% for this specific run. This slight difference arose because the $\epsilon=255/255$ evaluation batch was interrupted and had to be re-run. Since each run, including the restarted one, operated on a randomly drawn subset of 500 samples, minor variations in baseline accuracy due to sampling variability are statistically expected. The small difference observed (1.2 percentage points) falls well within the anticipated margin of error for this subset size, demonstrating the statistical consistency of the baseline measurements across runs despite the interruption. Consequently, the accuracy drop reported for the $\epsilon=255/255$ case is calculated relative to its specific 41.6\% baseline, while 42.8\% serves as the reference for all other perturbation levels.

The models were subjected to untargeted PGD attacks with $\epsilon$ values ranging from 2/255 (subtle) to 255/255 (maximum). The resulting VQA accuracies and accuracy drops are presented in two tables, separating subtle perturbations ($\epsilon \le 16/255$) from larger ones ($\epsilon \ge 128/255$).

Table \ref{tab:vqa_results_subtle} shows the performance under subtle perturbations, which are generally near-imperceptible. Table \ref{tab:vqa_results_large} shows the performance under larger, more perceptible perturbations.

\begin{table*}
    \centering
    \caption{VQA Accuracy (\%) under Subtle Adversarial Perturbations ($\epsilon \le 16/255$)}
    \label{tab:vqa_results_subtle}
    \resizebox{\textwidth}{!}{%
    \begin{tabular}{@{}lcccccc@{}}
    \toprule
     & Clean & \multicolumn{4}{c}{Adversarial Accuracy (Drop)} \\
    \cmidrule(lr){3-6}
    Model & Acc. (\%) & $\epsilon=2/255$ & $\epsilon=4/255$ & $\epsilon=8/255$ & $\epsilon=16/255$ \\
    \midrule
    LLaVA-1.5-13B           & 87.4 & 80.4 (–7.0) & 80.6 (–6.8) & 79.0 (–8.4) & 76.8 (–10.6) \\
    Llama 3.2 Vision-8B-2   & 42.8 & 36.2 (–6.6) & 32.4 (–10.4) & 33.0 (–9.8) & 36.2 (–6.6) \\
    \bottomrule
    \end{tabular}%
    }
\end{table*}

\begin{table*}
    \centering
    \caption{VQA Accuracy (\%) under Large Adversarial Perturbations ($\epsilon \ge 128/255$)}
    \label{tab:vqa_results_large}
    \resizebox{\textwidth}{!}{%
    \begin{tabular}{@{}lcccc@{}}
    \toprule
     & Clean & \multicolumn{2}{c}{Adversarial Accuracy (Drop)} \\
    \cmidrule(lr){3-4}
    Model & Acc. (\%) & $\epsilon = 128/255$ & $\epsilon = 255/255$ \\
    \midrule
    LLaVA-1.5-13B                     
        & 87.4 
        & 67.4 (–20.0)  
        & 51.4 (–36.0) \\
    Llama 3.2 Vision-8B-2 
        & 42.8
        & 37.4 (–5.4)   
        & 31.4\textsuperscript{\textdaggerdbl} (–10.2\textsuperscript{\textdaggerdbl}) \\
    \bottomrule
    \end{tabular}%
    }
    \vspace{1ex}
    \footnotesize{\textsuperscript{\textdaggerdbl}Clean accuracy was 41.6\% in the $\epsilon=255/255$ run. 
    }
\end{table*}

The tables show that for LLaVA, the accuracy drop increases steadily with $\epsilon$, reaching 36.0\% points at $\epsilon=255/255$. For Llama 3.2 Vision, the drop is significant at low $\epsilon$ (peaking at 10.4 points at $\epsilon=4/255$) but appears less sensitive to further increases in perturbation magnitude, with a drop of only 10.2 points at $\epsilon=255/255$ (relative to its 41.6\% baseline in that run).

\begin{figure*}
    \centering
    \includegraphics[width=0.9\textwidth]{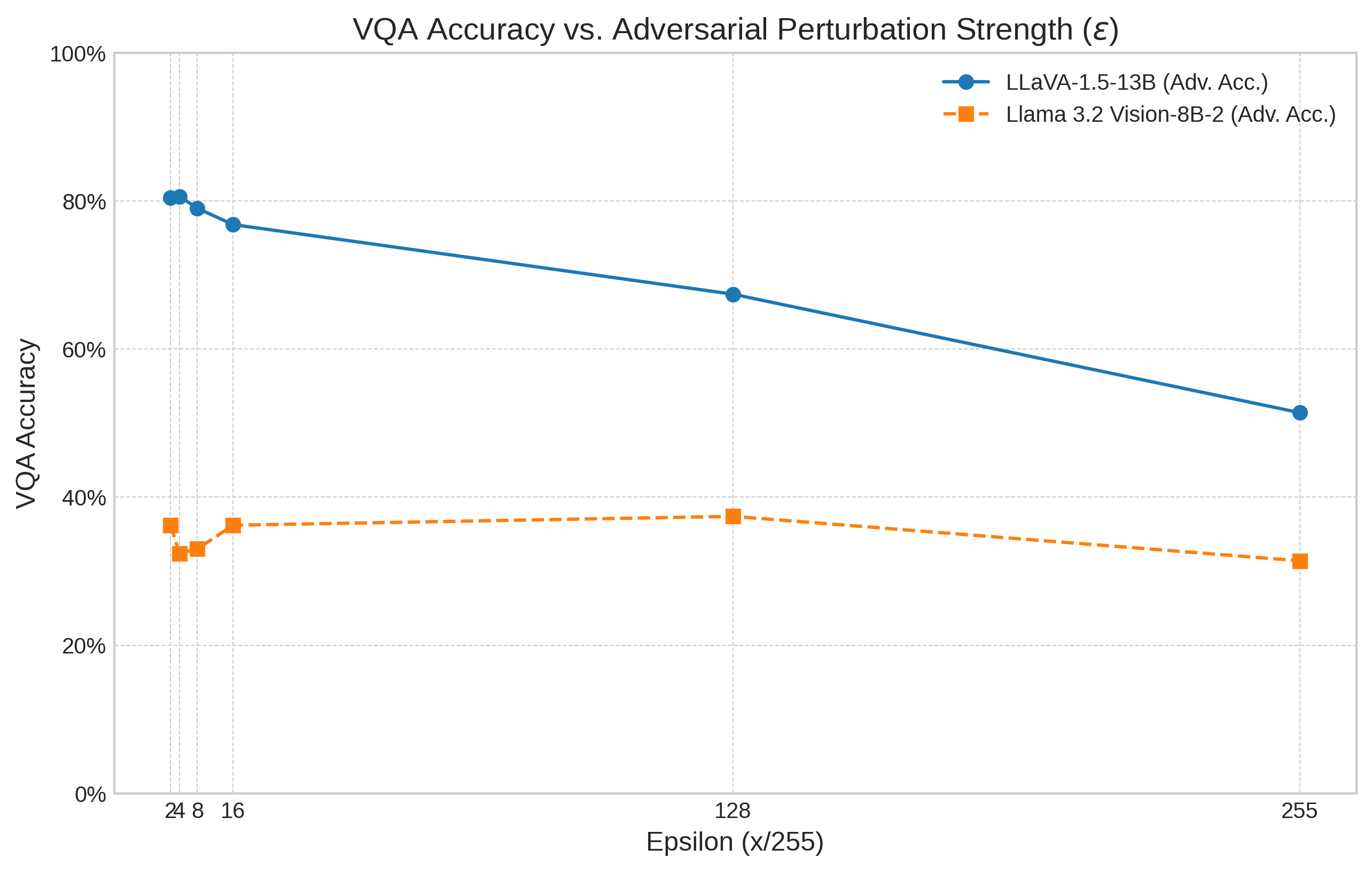}
    \caption{VQA Accuracy vs. Adversarial Perturbation Strength ($\epsilon$). Compares LLaVA-1.5-13B and Llama 3.2 Vision-8B-2 performance under PGD attack with varying $L_\infty$ budgets.}
    \label{fig:accuracy_drop}
\end{figure*}

Figure \ref{fig:accuracy_drop} visually represents the VQA accuracy for both models as the perturbation budget $\epsilon$ increases.

Qualitative analysis of individual samples revealed instances where adversarial perturbations caused models to generate incorrect, irrelevant, or nonsensical answers compared to their outputs on the corresponding clean images, particularly at higher perturbation levels.

\subsection{Comparative Analysis}
On the clean VQA v2 subset, LLaVA-1.5-13B demonstrated significantly higher baseline performance (87.4\% accuracy) compared to Llama 3.2 Vision-8B-2 (42.8\% accuracy). This substantial difference in clean accuracy highlighted the varying capabilities of the models on the VQA task prior to any adversarial manipulation.

When subjected to PGD attacks, the models exhibited distinct responses. LLaVA's accuracy degraded progressively as the perturbation budget ($\epsilon$) increased. The accuracy drop started at 7.0 percentage points for $\epsilon=2/255$ and reached a substantial 36.0 percentage points at the maximum perturbation of $\epsilon=255/255$.

In contrast, Llama 3.2 Vision showed a notable initial drop in accuracy even at low perturbation levels (e.g., 10.4 points at $\epsilon=4/255$), but its performance degradation appeared less sensitive to further increases in $\epsilon$. The maximum accuracy drop observed for Llama 3.2 Vision was 10.4 points (at $\epsilon=4/255$), and even at $\epsilon=255/255$, the drop was only 10.2 points relative to its baseline in that specific run.

Based purely on the magnitude of the accuracy drop under attack, Llama 3.2 Vision appeared quantitatively more robust, particularly at higher perturbation levels ($\epsilon \ge 128/255$). Its accuracy, while lower overall, degraded less severely in percentage points compared to LLaVA when strong adversarial noise was introduced. However, LLaVA maintained a higher absolute accuracy score even under the strongest attack ($\epsilon=255/255$, LLaVA accuracy 51.4\% vs Llama 3.2 Vision accuracy 31.4\%).

These differing robustness profiles might stem from several factors. LLaVA's simpler projection layer architecture could potentially be more susceptible to perturbations that directly affect the mapped visual features. Llama 3.2 Vision's more complex cross-attention adapter mechanism, combined with its significantly larger pre-training dataset (billions vs millions of image-text pairs) and potentially more sophisticated alignment process, might contribute to its relatively stable performance under increasing perturbation strengths, even though its baseline VQA performance was lower on this specific task and dataset subset. The larger vision encoder (ViT-H vs ViT-L) in Llama 3.2 Vision could also play a role, although previous work suggested encoder size alone might not guarantee robustness \citep{bhagwatkarImprovingAdversarialRobustness2024}.

\subsection{Critical Analysis}
The results provided insights into the adversarial robustness of Meta's Llama 3.2 Vision and its comparison with LLaVA. Addressing the first research question, Llama 3.2 Vision demonstrated vulnerability to PGD-based adversarial examples targeting its visual input. Its VQA accuracy dropped by up to 10.4 percentage points even with relatively subtle perturbations ($\epsilon=4/255$). This indicated that, despite its advanced architecture and extensive training, the model was not inherently immune to visually grounded adversarial attacks degrading its performance.

\subsubsection{Nuanced picture of robustness}
Regarding the second research question, comparing Llama 3.2 Vision to LLaVA revealed a nuanced picture of robustness. While LLaVA achieved significantly higher baseline VQA accuracy, it suffered a much larger degradation under attack, with accuracy dropping by up to 36.0\% points. Llama 3.2 Vision, despite its lower baseline, exhibited a smaller maximum accuracy drop (10.4 points) and greater stability against increasing perturbation strengths. Therefore, based on the relative decrease in performance under attack, Llama 3.2 Vision could be considered more robust than LLaVA in this specific experimental setup. The architectural differences (cross-attention adapter vs. simple projection) and the vastly larger scale of pre-training data used for Llama 3.2 Vision likely contributed to this greater resilience against stronger perturbations, potentially fostering more stable internal representations, even if not translating to superior baseline performance on this VQA subset.

These findings carry significant implications for the security of deployed VLMs. The observed vulnerability of both models, particularly Llama 3.2 Vision as a state-of-the-art open-weight model, underscores the risks associated with visual input channels. Even perturbations designed simply to maximise internal loss, without a specific malicious target, were sufficient to degrade performance noticeably. This suggests that real-world applications relying on these models could be susceptible to performance degradation or manipulation through adversarial images. The results also highlighted a potential trade-off: LLaVA's higher baseline performance came at the cost of lower robustness, while Llama 3.2 Vision's greater robustness was accompanied by lower baseline accuracy on this task. This suggests that optimising for standard benchmark performance might not automatically confer adversarial resilience.

\subsubsection{Limitations}
Several limitations should be acknowledged. Firstly, the evaluation was conducted on a subset of 500 samples from the VQA v2 dataset due to computational constraints; results on the full dataset or other datasets might differ. Secondly, the study focused exclusively on the PGD attack under an $L_\infty$ threat model. Other attack algorithms (e.g., CW attacks, AutoAttack) or norm constraints ($L_2, L_0$) could reveal different vulnerabilities. The attack was also untargeted; targeted attacks aiming for specific incorrect outputs might pose different challenges. Thirdly, robustness was evaluated solely on the VQA task. Performance degradation might vary on other multimodal tasks like image captioning or complex reasoning. Finally, while efforts were made to use appropriate hyperparameters, the computational cost limited exhaustive exploration of the PGD parameter space (iterations, step size).

\subsubsection{Adversarial attacks}
The findings align with broader research demonstrating the susceptibility of VLMs to adversarial attacks via their vision component \citep{lukebaileyImageHijacksAdversarial2023, schlarmannAdversarialRobustnessMultiModal2023, shayeganiJailbreakPiecesCompositional2023, dongHowRobustGoogles2023}. The vulnerability observed in Llama 3.2 Vision echoes the results from Meta's own evaluation of Llama Guard 3 Vision \citep{chiLlamaGuard32024}, where PGD attacks significantly impacted the safety classifier's performance, reinforcing the notion that visual perturbations pose a genuine threat even to models designed with safety considerations. While Bhagwatkar et al. \citep{bhagwatkarImprovingAdversarialRobustness2024} found a limited correlation between model/encoder scale and robustness in their study (which included LLaVA but not Llama 3.2 Vision), our results suggest that architectural choices (adapter vs. projection) and training scale/methodology might indeed influence relative robustness, warranting further investigation. The lower baseline but higher relative robustness of Llama 3.2 Vision compared to LLaVA presents an interesting data point in this ongoing discussion.

\section{Conclusions}

This paper has investigated the security of the vision component in contemporary open-source foundation models, specifically by evaluating the adversarial robustness of LLaVA-1.5-13B and Meta's Llama 3.2 Vision-8B-2. A further aim was to explore potential links between model architecture, training, and observed robustness. These aims were pursued through several objectives, which have been successfully met.

Both models were implemented, and their adversarial robustness was tested using untargeted PGD against the visual input modality. The experiments were conducted within an industry practice, formalised threat model, and empirically evaluated on the Visual Question Answering (VQA) v2 dataset subset.

The results of these adversarial attacks were quantified using the standard VQA accuracy metric, allowing for a direct comparison between the models' performance on clean versus perturbed images across various attack strengths ($\epsilon$). The evaluation compared the accuracy degradation (accuracy drop) of LLaVA and Llama 3.2 Vision, revealing distinct robustness profiles. Llama 3.2 Vision, despite a lower baseline accuracy in this setup, exhibited a smaller drop in performance under attack compared to LLaVA, particularly at higher perturbation levels. These quantitative findings were then evaluated in the context of the models' differing architectures (simple projection layer in LLaVA vs. cross-attention adapter in Llama 3.2 Vision) and their respective training scales and methodologies.

Through the successful completion of these objectives, the primary aim of evaluating and comparing the adversarial robustness of these two key open-weight VLMs was achieved, providing insights into the security of their visual components.

\subsection{Significant findings}
The empirical evaluation yielded several significant findings regarding the adversarial robustness of the vision component in the selected foundation models:

\begin{itemize}
    \item \textbf{Universal Vulnerability:} Both models demonstrated clear vulnerability to untargeted Projected Gradient Descent (PGD) attacks applied to the visual input. Measurable degradation in VQA accuracy occurred even with subtle, near-imperceptible perturbations ($\epsilon \le 16/255$).
    \item \textbf{Distinct Robustness Profiles:} The models exhibited markedly different responses to increasing perturbation strengths. LLaVA's accuracy degraded progressively and substantially with larger $\epsilon$, suffering a maximum drop of 36.0 percentage points. Conversely, Llama 3.2 Vision showed a significant initial drop but greater stability against stronger perturbations, with its accuracy drop plateauing around 10 percentage points. However, LLaVA maintained higher absolute accuracy scores across all perturbation levels due to its superior baseline performance - but this was likely due to Llama 3.2 implementation.
    \item \textbf{Architectural and Training Implications:} The observed difference in relative robustness tentatively suggests that architectural choices (Llama 3.2 Vision's cross-attention adapter vs. LLaVA's simpler projection) and the scale of pre-training data may influence a model's resilience to visual adversarial examples, potentially favouring more complex integration mechanisms and larger datasets for stability, even if not for peak task performance in all settings.
\end{itemize}

Collectively, these findings confirm that the vision modality represents a viable attack vector for degrading the performance of contemporary open-weight VLMs, including Meta's Llama 3.2 Vision. Furthermore, they highlight that adversarial robustness does not necessarily correlate directly with standard benchmark performance and may be influenced by underlying architectural and training factors.

Overall, we believe that this paper is one of the first to provide systematic comparisons of visual adversarial robustness in popular open models, and outlines that robustness does not always align with standard accuracy metrics.

\subsection{Future Work}
Building upon the findings and acknowledging the limitations of this study, several avenues for future research emerge. This includes the replicating the experiments on either the full VQA v2 dataset or a larger subset, and other multimodal benchmarks would provide a more comprehensive assessment of robustness across different tasks and data distributions. Along with this, there is the opportunity for diverse adversarial attacks, such as investigating vulnerability to a wider range of attack algorithms beyond untargeted PGD, like Carlini \& Wagner (CW) attacks or targeted attacks aiming to induce specific incorrect outputs (e.g., Image Hijacks), would offer a more complete security assessment. 

In relation to investigating Performance Discrepancies, further work is needed to pinpoint the exact reasons for the observed lower baseline performance of Llama 3.2 Vision on the VQA subset compared to reported benchmarks, potentially involving different prompt configurations or preprocessing steps. With architecture and training Analysis, a deeper analysis into how specific components (e.g., the design of multimodal adapters) and training phases (pre-training data scale, alignment techniques like RLHF) quantitatively impact adversarial robustness would be highly valuable. With defence mechanisms, research into developing and evaluating effective defence strategies specifically tailored for VLMs like Llama 3.2 Vision is essential for mitigating the identified risks. 

Finally, there is a need to further investigate native multimodality vs. adapters. This would involve investigating newer architectures, such as Meta's Llama 4, which reportedly employs early fusion to jointly pre-train text and image data, and could reveal differences in robustness compared to the adapter-based approach of Llama 3.2 Vision. Understanding how native multimodal training impacts vulnerability is highly relevant for the field.

\bibliography{sn-bibliography}

\end{document}